\documentclass[11pt,a4paper]{article}
\usepackage[]{acl2021}
\usepackage{times}
\usepackage{latexsym}

\usepackage{microtype}
\usepackage{subcaption}
\usepackage{multirow}
\usepackage{booktabs}
\usepackage{graphicx}
\usepackage{amsmath,amssymb,bm}

\title{Human-in-the-loop Abstractive Dialogue Summarization}

\author{
 Jiaao Chen$^{\dagger}$ \quad Mohan Dodda$^{\dagger}$  \quad Diyi Yang$^\diamond$ \\
 $^\dagger$ Georgia Institute of Technology \quad $^\diamond$ Stanford University \\
\texttt{\{jchen896,mohandodda\}@gatech.edu} \\ 
\texttt{ diyiy@cs.stanford.edu }
}

\date{}

\begin{document}
\maketitle
\begin{abstract}
Abstractive dialogue summarization has received increasing attention recently. Despite the fact that most of the current dialogue summarization systems are trained to maximize the likelihood of human-written summaries and have achieved significant results, there is still a huge gap in generating high-quality summaries as determined by humans, such as coherence and faithfulness, partly due to the misalignment in maximizing a single human-written summary. To this end, we propose to incorporate different levels of human feedback into the training process. This will enable us to guide the models to capture the behaviors humans care about for summaries. Specifically, we ask humans to highlight the salient information to be included in summaries to provide the \textit{local feedback}, and to make overall comparisons among summaries in terms of coherence, accuracy, coverage, concise and overall quality, as the \textit{global feedback}. We then combine both local and global feedback to fine-tune the dialog summarization policy with Reinforcement Learning. Experiments conducted on multiple datasets demonstrate the effectiveness and generalization of our methods over the state-of-the-art supervised baselines, especially in terms of human judgments. 
\end{abstract}

\section{Introduction}
Abstractive conversation summarization, which aims at processing, organizing, and distilling human interaction activities into natural, concise, and informative text \cite{murray-etal-2006-incorporating,wang2013domain}, is one of the most challenging and interesting tasks in text summarization. Growing attention has been paid to neural abstractive conversation summarization through a variety of designs including transferring document summarization models \cite{gliwa-etal-2019-samsum,yu2021adaptsum,https://doi.org/10.48550/arxiv.2204.13498}, utilizing conversational structures \cite{chen-yang-2020-multi,feng2020incorporating,Zhu_2020,chen2021structureaware,liu2019fast,https://doi.org/10.48550/arxiv.2205.13190,zhang-etal-2022-focus, https://doi.org/10.48550/arxiv.2106.08556}, introducing conversational data augmentation \cite{chen-yang-2021-simple}, incorporating controllable signals \cite{https://doi.org/10.48550/arxiv.2104.07606, https://doi.org/10.48550/arxiv.2105.14064} and pre-training conversation models \cite{https://doi.org/10.48550/arxiv.2109.02492}. 
Most of them are trained with supervised learning, which maximizes the log probability of human written summaries. While they have gained impressive performances, there are still huge gaps in generating high-quality summaries as determined by humans such as coherence or faithfulness\cite{chen2021structureaware}, largely due to a misalignment between the fine-tuning objective (\textit{maximizing the likelihood of single human-written summary}) and the actual needs (\textit{generating more human-favored summaries}) \cite{https://doi.org/10.48550/arxiv.1909.08593}. 

To train the summarization models on objectives that can more closely capture the behaviors humans care about, Reinforcement Learning (RL) has been used to directly optimize the rewards learned and constructed from human feedback \cite{https://doi.org/10.48550/arxiv.1909.08593,https://doi.org/10.48550/arxiv.2009.01325,https://doi.org/10.48550/arxiv.1909.01214,ye-simpson-2021-proposal}. Different kinds of feedback have been explored to construct the reward functions such as human ratings over CNN/DM summaries \cite{https://doi.org/10.48550/arxiv.1909.01214}, overall preferences among pairs of summaries \cite{https://doi.org/10.48550/arxiv.1909.08593}, and the similarity-redundancy matrix \cite{https://doi.org/10.48550/arxiv.1909.01214}. While achieving promising performances, they are mainly designed for document summarization with a single reward function learned from overall assessments on summaries\cite{https://doi.org/10.48550/arxiv.1909.01214,https://doi.org/10.48550/arxiv.1909.08593}. As a result, they might not be directly adapted to dialogue summarization because of the intrinsic differences between documents and conversations. Compared to documents, conversations are generally less structured and more complex \cite{chen-yang-2020-multi}. There are diverse interactions between multiple speakers and complex structures such as interruptions, discourse relations, and speaker roles in dialogues \cite{chen-yang-2020-multi}. Therefore, more subtle levels of human feedback with the consideration of \textit{conversation structural information} is needed to provide more comprehensive, consistent, and generalizable rewards, which may lead to better performances for dialogue summarization.

To fill in this gap, we introduce Human-In-The-Loop (HILT) abstractive dialogue summarization with different levels of human feedback to leverage various conversation structures. Specifically, we incorporate two levels of human feedback: (1) \textbf{Local} Feedback, which consists of highlighted words or phrases in dialogues to capture the salient structural information, including but not limited to \textit{speaker's intents}, \textit{identifiable events/topics},  and \textit{discourse relations} (e.g., causal relationships and \textit{important emotions}), and (2) \textbf{Global} Feedback, which includes dimensions like \textit{Coherence}, \textit{Accuracy}, \textit{Coverage}, \textit{Concise} and the \textit{Overall Quality},  to provide more comprehensive human preferences on the given summary. We hire and train human annotators to provide the introduced two levels of human feedback on 1,000 randomly sampled conversations from the DialogSum dataset \cite{chen-etal-2019-bag}. With the collected human feedback, we construct the \textbf{local reward ($r_l$)} based on the similarities between the generated summaries and the annotated highlights and learn the \textbf{global reward ($r_g$)} models via supervised learning which predict the human preferences. Finally, we train the summarization policy via RL to maximize the rewards predicted by $r_l$ and $r_g$. Specifically, the policy generates a token of text at each time step and is updated using the PPO algorithm \cite{https://doi.org/10.48550/arxiv.1909.08593} based on the reward given to the entire generated summary. We conducted extensive experiments and ablation studies in different settings on the recent conversation summarization dataset, DialogSum \cite{chen-etal-2019-bag} and SAMSum \cite{gliwa-etal-2019-samsum}, to demonstrate the superiority of our methods compared to the state-of-the-art supervised learning baselines, especially in terms of human judgments and generalization abilities.

To summarize, our contributions are: (1) we introduced and collected the local and global feedback tailored for abstractive conversation summarization; (2) we designed the HITL to learn better conversation summarization policies via reinforcement learning where different levels of human feedback are directly optimized; (3) we performed extensive experiments to study the effectiveness and generation abilities of our HITL methods on DialogSum and SAMSum datasets.

\begin{figure*}[ht]
\centering
\includegraphics[width=2.0\columnwidth]{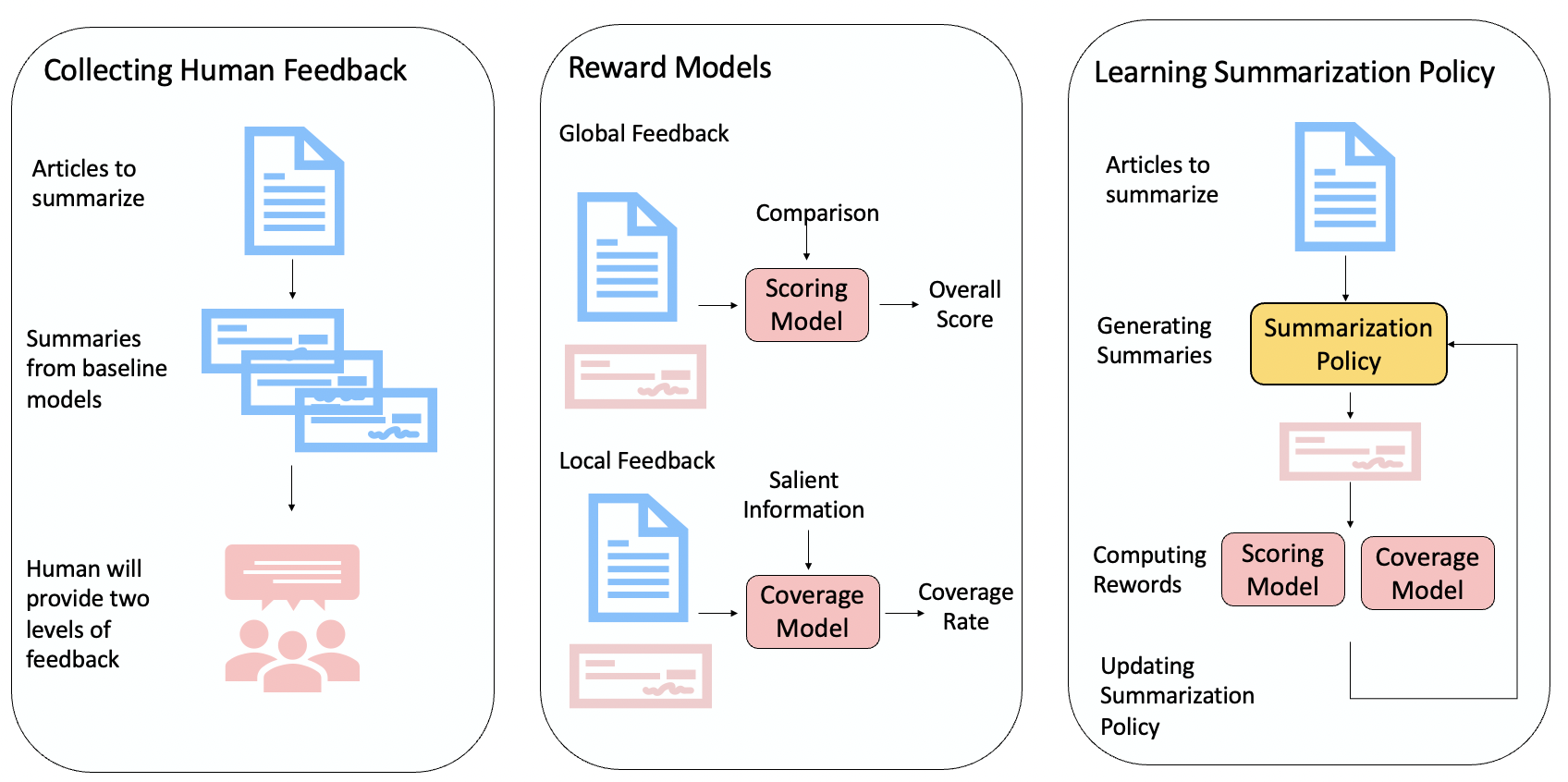}
\caption{Overall process of our human-in-the-loop conversation summarization system including collecting human feedback, learning and designing reward models based on feedback, and learning the summarization policy.}
\label{Fig:model}
\end{figure*}

\section{Related Work}
\subsection{Abstractive Dialogue Summarization}
Neural abstractive dialogue summarization has received intensive attention recently with the introduction of large scale datasets \cite{gliwa-etal-2019-samsum,chen-etal-2019-bag,tuggener-etal-2021-summarizing}. Besides directly transferring documents summarization methods to conversations \cite{gliwa-etal-2019-samsum}, models tailored for conversation have been proposed to achieve better performances \cite{10.1145/3308558.3313619, zhu2020hierarchical, https://doi.org/10.48550/arxiv.2107.03175}, which make use of the rich structured information in conversations such as dialogue acts \cite{Goo_2018}, key point/entity sequences \cite{10.1145/3292500.3330683,narayan2021planning},  topic segments \cite{Liu_2019, li-etal-2019-keep}, stage developments \cite{chen-yang-2020-multi}, discourse relations \cite{chen2021structureaware,feng2020dialogue}, action mentions \cite{zhang-etal-2022-focus}, and coreferences  \cite{https://doi.org/10.48550/arxiv.2106.08556}. Recent work has also explored learning in a data efficient way through data augmentation and semi-supervised learning \cite{chen-yang-2021-simple}, generating more controllable summaries \cite{https://doi.org/10.48550/arxiv.2105.14064,https://doi.org/10.48550/arxiv.2104.07606}. Moreover, external information such as commonsense knowledge is incorporated to help understand the global conversation context \cite{feng2020incorporating}. \citet{https://doi.org/10.48550/arxiv.2109.02492} pre-trained a language model on conversational data to help the summarization as well.

Most of the current dialogue summarization systems are still trained to maximize the likelihood of human-written text and have led to significant performances, but there is still a huge gap from generating high-quality summaries as determined by humans such as coherence, faithfulness, conciseness, and concreteness \cite{chen-yang-2020-multi}. This is mainly due to the misalignment between training objective and model evaluation. For example, models never plan and look ahead for overall summarization goals. To fill in this gap, we directly learn the summarization policy that maximizes the rewards constructed from human feedback via Reinforcement Learning to generate more human-favored summaries.

\subsection{Learning with Human Feedback}
Recent research has started to explore incorporating human feedback into the training process to achieve human-preferred systems in different tasks such as dialogue generation \cite{https://doi.org/10.48550/arxiv.1907.00456,Yi2019TowardsCA,https://doi.org/10.48550/arxiv.1901.05415}, story generation \cite{https://doi.org/10.48550/arxiv.2002.05058}, document summarization \cite{https://doi.org/10.48550/arxiv.1909.08593,https://doi.org/10.48550/arxiv.2009.01325,https://doi.org/10.48550/arxiv.1909.01214} and etc. Our work is most related to previous work which utilizes human feedback to train document summarization models with Reinforcement Learning (RL) \cite{https://doi.org/10.48550/arxiv.1909.08593,https://doi.org/10.48550/arxiv.2009.01325,https://doi.org/10.48550/arxiv.1909.01214,ye-simpson-2021-proposal}, where human ratings/comparisons over summaries are usually used to learn the reward models to serve as the value networks in RL. Despite the effectiveness, it is challenging to directly apply them to conversation summarization, largely due to the complex structures in conversations, which requires subtle reward design. 

Inspired by these prior work, we introduce two levels of human feedback to guide the dialogue summarization model to generate more human-favored summaries, including the (1) \textbf{Local} Feedback which highlights the important conversation structures to summarize and (2) the \textbf{Global} Feedback which consists of different dimensions to provide more comprehensive judgments. 

Our work is also related to using RL to optimize automatic metrics for summarization, such as ROUGE \cite{https://doi.org/10.48550/arxiv.1511.06732,https://doi.org/10.48550/arxiv.1804.07036,https://doi.org/10.48550/arxiv.1907.12894,https://doi.org/10.48550/arxiv.2106.04080}, while we are directly optimizing human preferences with RL.

\section{Methods}
In this section, we introduce our \textbf{H}uman-\textbf{i}n-\textbf{t}he-\textbf{L}oop conversation summarization (\textbf{HITL}) pipeline where we incorporate two levels of human feedback, the \textbf{local} and \textbf{global} feedback, into the learning process. Inspired by \citet{https://doi.org/10.48550/arxiv.2009.01325}, our pipeline for abstractive conversation summarization includes 3 stages: (1) Collecting two levels of human feedback from conversation-summary pairs where summaries are generated with baseline models; (2) Learning and designing reward models from two levels of human feedback; (3) Learning the summarization policy which could generate higher-quality summaries as judged by humans against the reward model. We visualize the overall process in Figure~\ref{Fig:model}.

\subsection{Datasets}
We utilize DialogSum \cite{chen-etal-2019-bag}, a recent large scale dialogue summarization dataset with the emphasis on real life daily conversations, to study the human-in-the-loop conversation summarization. We selected DialogSum because the summaries in DialogSum are less extractive with more novel n-grams and the summaries are more compressed compared to the other conversation summarization datasets \cite{chen-etal-2019-bag}\footnote{The data statistics are shown in Table~\ref{Tab:data_stats} in the Appendix.}, which makes the datasets more challenging and requires human knowledge to generate better summaries. 

\subsection{Collecting Human Feedback} 
Here we describe the process on getting the desired global and local human feedback. Basically, for every dialogue, we would ask human annotators to highlight the important information to serve as the local feedback (Section~\ref{Sec:Local}). For every dialogue, we will first generate 4 baseline summaries and then ask human annotators to make comparison among them to get our global feedback (Section~\ref{Sec:Global}).

\begin{table}[t]
\center
\small
\begin{tabular}{cccc}
\toprule
\textbf{Dataset} &\textbf{\# Turns}   &\textbf{\# Words}  &\textbf{\# Words in Sum} \\ \midrule \midrule
Sampled &9.6 &127.6 &22.9    \\ \midrule
DialogSum &9.8 &131.0 & 23.6 \\
  \bottomrule             \end{tabular} \caption{Data statistics of sampled 1000 dialogues and DialogSum including the average number of turns and words in conversations and the average number of words in ground truth summaries. } \label{Tab:stats}
\end{table}

\subsubsection{Annotation Setup}
\paragraph{Sampling Dialogues} From this DialogSum dataset, we randomly sample 1,000 dialogues from 13,360 dialogues to collect our designed two levels of human feedback. As the data statistics shown in Table~\ref{Tab:stats}, the distribution of our sampled examples is close to that of the orignal DialogSum dataset.

\paragraph{Baseline Summaries} We generate a set of baseline summaries with different models for the global feedback annotation. Specifically, for every dialogue, we generate 4 summaries with 4 different summarization systems: (1) BART-large fine-tuned on SAMSum and XSUM \footnote{\url{https://huggingface.co/Salesforce/bart-large-xsum-samsum}} with a 30.4/11.5/24.8 ROUGE score on DialogSum, (2) DistilBART fine-tuned on CNN/Daily Mail and SaumSUM \footnote{\url{https://huggingface.co/philschmid/distilbart-cnn-12-6-samsum}} with a 33.8/13.6/27.8 ROUGE score , (3) BART-large fine-tuned on SAMSum \footnote{\url{https://huggingface.co/linydub/bart-large-samsum}} and with a 33.0/13.5/27.0 ROUGE score (4) BART-large-xsum \footnote{\url{https://huggingface.co/facebook/bart-large-xsum}} fine-tuned on SAMSum \footnote{\url{https://huggingface.co/knkarthick/meeting-summary-samsum}} with a 26.6/10.2/21.4 ROUGE score. These different summaries are then compared by human to provide global feedback.

\paragraph{Hiring and Training Annotators}
We hire two annotators through Upwork\footnote{\url{https://www.upwork.com}} and provide them with extensive training for the task. During multiple training sessions, we explain how to highlight salient information and compare summaries using our interfaces. We go through selected example dialogues and discuss with them to resolve inconsistencies and disagreements. To further reaffirm the training, we also perform test runs on the sampled dialogues. From these test cases, we make sure that they annotate the data properly and achieve good agreements. We pay the annotators \$25 per hour. We get 41.67 hours of work for the first member and 39.83 hours for the second member \footnote{The interface is shown in the Appendix}. 

\subsubsection{Local Feedback} \label{Sec:Local}
For the local feedback, we ask annotators to highlight the salient information in the provided dialogues. The highlighted information needs to be helpful in generating a summary. The information can be phrases, sentences, or a couple of words in the given dialogues. Specifically, we ask annotators to look for some important aspects including (1) \textit{speaker's intents}, (2) \textit{identifiable events/topics}, (3) \textit{discourse relations} such as causal relationships and (4) important \textit{emotions} in the conversation. For every conversation, we ask the annotator to annotate 3 to 8 highlights. After 3 rounds of training sessions, we examine the quality by asking them to annotate the same set of 50 dialogues and computing the agreement scores between the two annotators (0.865 BERT-score between their annotated spans) \footnote{The BERTscore for randomly sampled pairs of spans is 0.573.}. We also make sure the highlights match the important information in ground truth summaries (0.792 BERT-score between annotated spans and corresponding summaries) \footnote{The BERTscore for randomly sampled pairs of utterance and summary is 0.469.}.  Annotators then annotate the remaining dialogues by themselves independently. After annotation, we collect 6.1 spans for every dialogue with 59.5 words on average.

\subsubsection{Global Feedback} \label{Sec:Global}
After highlighting the salient information, we provide the annotators with 3 pairs of summaries sampled from the set of baseline summaries. We then ask them to make comparisons in terms of \textit{Coherence, Accuracy, Coverage, Conciseness}, and \textit{Overall Quality}. For every comparison between summary A and summary B, the annotators need to grade upon a scale of 5 points: the summary A mostly better, summary A partially better, equal, summary B partially better, the summary B mostly better. We provide detailed guidelines to the annotators about those different dimensions\footnote{The detailed guidelines are shown in the Appendix}.  After 3 rounds of training sessions, we show the annotators 50 dialogues with 150 pairs of summaries and ask them to make comparisons, resulting in 150 comparisons. We then calculate the Fleiss Kappa scores to measure the agreements among different annotators.  We obtain an average score of 0.342 for Coverage, 0.381 for Coherence, 0.376 for Conciseness, 0.373 for Accuray, and 0.369 for Overall Quality, indicating moderate agreement \cite{fleisskappa}. Annotators then annotate the remaining dialogues by themselves independently. In total, we collect 3000 pairs of comparisons for every dimension.

\begin{table*}[t]
\center
\begin{tabular}{c|c|c|c|c|c}
\toprule
\textbf{Methods}        & \textbf{\# Training Data} & \textbf{Rewards} & \multicolumn{1}{l}{\textbf{ROUGE-1}}       & \multicolumn{1}{l}{\textbf{ROUGE-2} }       & \multicolumn{1}{l}{\textbf{ROUGE-L}}       \\ \midrule \midrule
BART-large     & Full        & -       & 47.28 & 21.18 & 44.83 \\  \midrule 
HITL-synthesis & Full        & $r_g$      & 46.87 & 21.03 & 45.12 \\
HITL-synthesis & Full        & $r_l$      & 47.27 & 22.18 & 45.15 \\
HITL-synthesis & Full        & $r_g+r_l$   & 47.46 & 22.13 & 45.24 \\ \midrule
HITL-synthesis & 1000        & $r_g$      & 46.25 & 20.79 & 44.37 \\
HITL-synthesis & 1000        & $r_l$      & 46.18 & 21.12 & 45.13 \\
HITL-synthesis & 1000        & $r_g+r_l$   & 46.38 & 21.26 & 45.08 \\ \midrule 
HITL$\dag$         & 1000        & $r_g$       & 47.54                         & 23.05                         & 45.38                         \\
HITL$\dag$          & 1000        & $r_l$       & 47.88                         & 23.17                         & 45.87                         \\
HITL$\dag$           & 1000        & $r_g+r_l$     & \textbf{48.29}                & \textbf{23.65}                & \textbf{46.23}   \\ \bottomrule             
\end{tabular} \caption{ ROUGE-1, ROUGE-2 and ROUGE-L scores for different models on the DialogSum Corpus test set. $\dag$ means our model. We performed Pitman's permutation test \cite{dror2018hitchhiker} and found that \emph{HITL} significantly outperformed the supervised baseline \emph{BART-large} ($p < 0.05$). The results are averaged over three random runs.}\label{Tab: rouge_results} 
\end{table*}

\subsection{Method}
This section focuses on how to incorporate the annotated feedback into the training process to assist the summarization systems in generating more human-favored summaries.

\subsubsection{Rewards Modeling}
We first describe how to train the reward models and compute the rewards for any given conversation-summary pairs based on the collected human feedback. 

\paragraph{Local Rewards}
Our goal is to encourage the summarization systems to generate summaries that cover the important information mentioned in the dialogues while avoiding redundant information. Thus here we propose to model the local rewards based on these highlights from annotators. For a given conversation $C$ with a set of human-annotated salient spans $M = M_{1:m}$ (e.g., phrases/sentences/words in the dialogues), suppose the model would generate a summary $s$.  We view the list of highlights $M$ annotated by humans as information needed by the summaries, and the other sentences without highlights as possible redundant information set $N = N_{1:n} = C - M$. We then calculate the local coverage rewards $r_l(C, s, M)$ by calculating the cosine distances between the embeddings of the summary and the information in the dialogues:
\begin{align}
    r_l(C, s, M) = \sum_i^m cos(s, M_i) - \sum_j^n cos(s, N_j)
\end{align}

Here we embed the summaries and the dialogue information utilizing sentence-transformers (all-mpnet-base-v2) \footnote{\url{https://github.com/UKPLab/sentence-transformers}} \citep{reimers-gurevych-2019-sentence}.

\paragraph{Global Rewards}
Generating high-quality summaries with better human preferences is essential for building better summarization systems. To this end,  we design the global rewards by learning human preferences from their annotations. For a given set of annotated conversations $C = \{C_1, ..., C_n\}$ with baseline summaries $S = \{(s_1^1, s_2^1, s_3^1, s_4^1), ..., (s_1^n, s_2^n, s_2^n, s_3^n)\}$ with different dimensions of global human feedback, we first learn a set of reward models $r_{g_j}(C, s;\theta_e, \theta_j)$ to measures the quality or impact of the dimension $j$ on summary $s$ for a conversation $C$. Here,  $\theta_e$ are the parameters to encode the conversation and summaries; $\theta_j$ stands for the parameters of linear heads for the dimension $j$, which outputs a scalar on top of $\theta_e$. Specifically, we initialize $\theta_e$ with a BART-large model fine-tuned on the DialogSum dataset, and randomly initialize $\theta_j$ for every dimension in the global feedback. During training, we train the model to predict which summary in a summary pair $\{s_n^i, s_m^i\}$ of conversation $C_i$ is preferred by human, by optimizing the following objective:
\begin{align*}
    \mathcal{L} = - &\mathbb{E}_{(C_i, s_n^i, s_m^i) \sim (C,S)} \Sigma_j [\log (\sigma (r_{g_j}(C_i, s_n^i; \\
    & \theta_e, \theta_j) - r_{g_j}(C_i, s_m^i;\theta_e, \theta_j)))]
\end{align*}
where $s_n^i$ is the summary preferred by humans. 

Implementations are shown in Section~\ref{Sec:implementation}. We select the hyper-parameter based on the loss on the validation set (8:2 split), and further evaluate the learned reward models in Section~\ref{Sec:human}.

We then combine different dimensions to provide the global rewards $r_g(C,s)$:
\begin{align}
    r_g(C,s) = \Sigma_j r_{g_j}
\end{align}

\begin{table}[t]
\center
\begin{tabular}{c|c}
\toprule
\textbf{Methods}        & \textbf{Human Preferred \%}  \\ \midrule \midrule
BART-large     & 18\%   \\   
HITL-($r_g+r_l$) $\dag$          & \textbf{82\%}  \\  \bottomrule             \end{tabular} \caption{Human preferences when comparing summaries generated by supervised baseline model (BART-large) and our best HITL model ($r_g+r_l$). $\dag$ means our method.} \label{Tab: human_results}
\end{table}

\subsubsection{HITL Summarization Policy Learning}
Here we train a summarization policy with human feedback for generating high-quality outputs as judged by humans.  We utilize reinforcement learning to learn the summarization policy $\pi_{\phi}^{\mathrm{RL}}$. Specifically, we initialize the policy with a supervised learning BART-large baseline model $\pi^{\mathrm{B}}$ fine-tuned on DialogSum. We use the PPO algorithm \cite{https://doi.org/10.48550/arxiv.1707.06347} to maximize the rewards from the above local and global reward models $r_l$ and $r_g$, where each time step is a BPE token \footnote{The reward model would give the rewards after the entire summary generated. Each episode terminates when the policy outputs the EOS token, and the discount factor $\gamma = 1$.}. The full reward $R(C,s, M)$ is:
\begin{equation}
    \begin{aligned}
    R(C,s, M) &= 
    w_l r_l(C, s, M) +  w_g r_g(C,s) \\
    &-\beta \log \left[\frac{\pi_{\phi}^{\mathrm{RL}}(s|C)}{\pi^{\mathrm{B}}(s|C)}\right]
    \end{aligned}
\end{equation}
We introduce a KL divergence term between the HITL policy and the supervised baseline model \cite{https://doi.org/10.48550/arxiv.1907.00456}. This term could prevent the learned policy generating outputs that are too different from the supervised  models and encourage the learned policy to explore instead of collapsing to a single model \cite{https://doi.org/10.48550/arxiv.2009.01325}. $w_l$, $w_g$ and $\beta$ are weights to balance different sub-rewards.

Following \citet{https://doi.org/10.48550/arxiv.2009.01325}, we use a Transformer with separate parameters from the policy for the PPO value function. And we initialize the value function to the parameters of
the reward model.

\begin{table}[t]
\center
\begin{tabular}{c|c}
\toprule
\textbf{Metric}        & \textbf{Agree with Human \%}  \\ \midrule \midrule
 ROUGE    & 55.3\%   \\   \midrule
 Coherence           & 62.4\%  \\  
 Accuracy           & 56.8\%  \\  
 Coverage           & 63.6\%  \\  
 Concise           & 59.5\%  \\  
 Over Quality           & 65.5\%  \\  \midrule 
 $r_g$           & \textbf{69.8\%}  \\  \bottomrule    \end{tabular} \caption{Agreement with human preferences of different reward models including the ROUGE score, coherence reward model, accuracy reward model, coverage  reward model, concise  reward model, over quality reward model and our global reward model $r_g$.} \label{Tab:reward}
\end{table}

\begin{table*}[t]
\center
\begin{tabular}{c|c|c|c|c|c} \toprule
\textbf{Methods} & \textbf{Training Data} & \textbf{Transferred Parts} & \textbf{ROUGE-1} & \textbf{ROUGE-2} & \textbf{ROUGE-L} \\ \midrule \midrule
BART-large                  & DialogSum & Whole Model     & 31.74                       & 5.93                        & 29.79                       \\ \midrule
HITL-($r_g+r_l$)$\dag$       & DialogSum  & Whole Model   & \textbf{33.58}             & \textbf{7.84}               & \textbf{32.63}   \\ \midrule \midrule

BART-large                  & SAMSum   & Whole Model     & 53.12                       & 27.95                       & 49.15                       \\ \midrule
HITL-($r_g$) $\dag$ $\dag$         & SAMSum & Global Reawrds    & \textbf{53.76}              & \textbf{28.04}               & \textbf{50.56}   \\  

\bottomrule      
\end{tabular} \caption{ROUGE-1, ROUGE-2 and ROUGE-L scores on the \textit{SAMSum test data}. $\dag$ means that we directly apply our HITL models trained on DialogSum to SAMSum. $\dag$ $\dag$ means that we re-train the policy on SAMSum corpus with the global reward learned from DialogSum annotations. The results are averaged over three random runs.} \label{Tab:zero_shot}
\end{table*}

\begin{table}[t]
\center
\begin{tabular}{c|c|c|c} \toprule
 \textbf{Quality} & \textbf{R1} &\textbf{R2} &\textbf{RL} \\ \midrule \midrule
                  Synthesis        & 46.38 &21.26 &45.08                       \\ 
        Noisy     & 46.32 &21.38 &44.76  \\ 
       Clean     & \textbf{47.58} &\textbf{22.58} &\textbf{45.56}   \\ \bottomrule           
\end{tabular} \caption{ROUGE scores on the DialogSum test data where the HITL-($r_g+r_l$) policy is learned with 400 annotations with different qualities. The results are averaged over three random runs.} \label{Tab:nosiy}
\end{table}


\begin{figure}[ht]
\centering
\includegraphics[width=0.9\columnwidth]{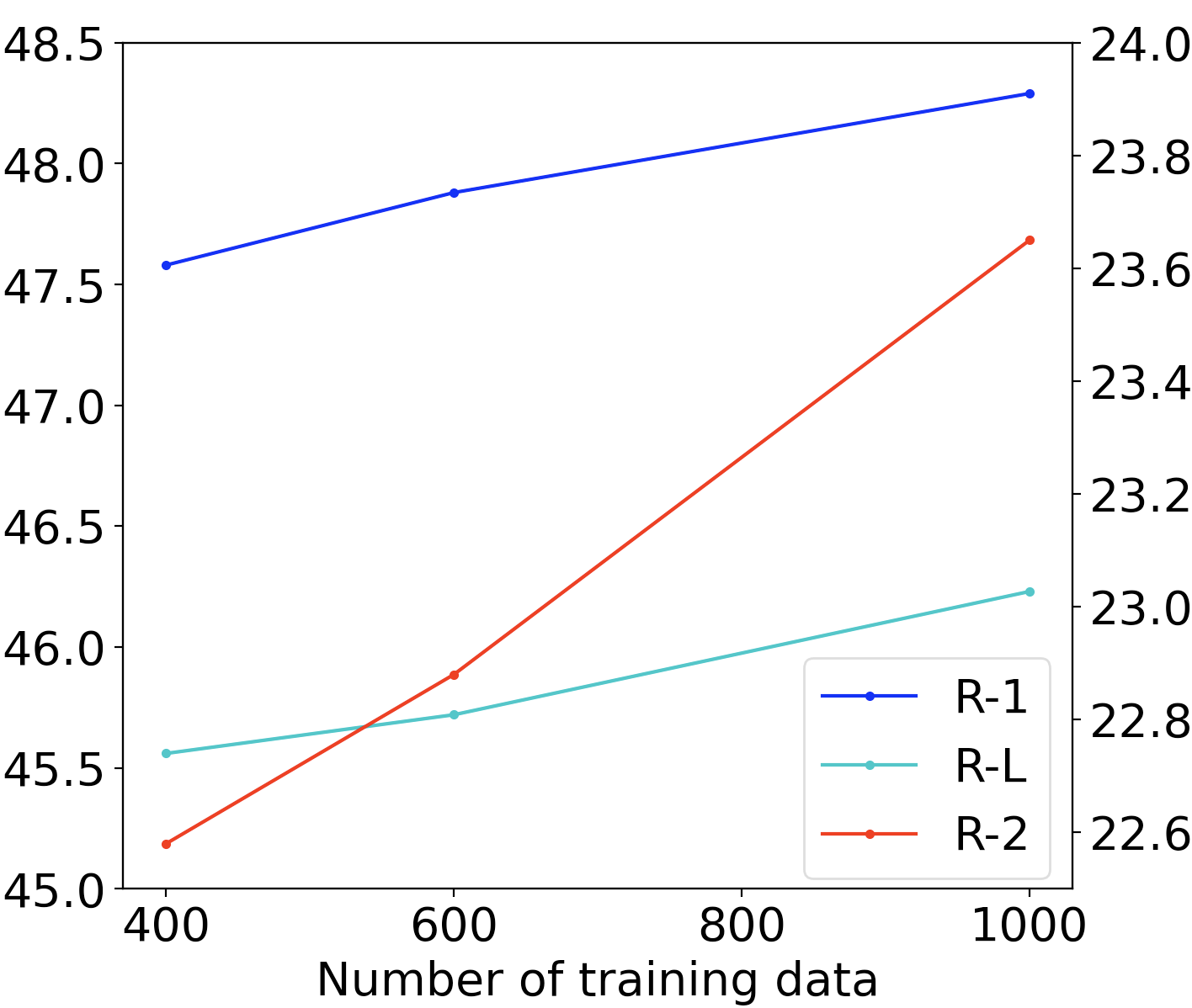}
\caption{ROUGE-1, ROUGE-2 and ROUGE-L scores on the DialogSum test data where the HITL policy is learned with different number of annotations (400, 600, 1000). The left y axis means ROUGE-1 and ROUGE-L, the right y axis means ROUGE-2.}
\label{Fig:nnumber}
\end{figure}

\section{Experiments}

\subsection{Baselines} \label{Sec:baseline}
We compare our models with several baselines:
\begin{itemize}
    \item \textbf{BART-large} \cite{lewis-etal-2020-bart}: We utilized BART-large as our backbone model as well as the supervised learning baseline. Utterances are separated by a special token.
    \item \textbf{HILT-synthesis}: We use heuristics to approximate the local and global feedback, via which we then learn synthesized reward models and the HITL summarization policy. Specifically, for the local feedback, we utilize a greedy algorithm \cite{https://doi.org/10.48550/arxiv.1611.04230, zhang-etal-2022-focus} to obtain the synthesis highlights based on ground truth summaries. For the global feedback, we utilize the randomly sampled utterance as negative summaries compared to the ground truth summary. 
\end{itemize}

\subsection{Implementation Details} \label{Sec:implementation}
For the supervised baseline, we initialize the model with BART-large and fine-tune it on the full DialogSum for 10 epochs with a 3e-5 learning rate and 120 warm-up steps. We use a batch size of 8. For the global reward models, we set the hidden size of the linear head 256. We use a batch size of 8 and train the reward model for 2 epochs with a 3e-5 learning rate. For PPO, we initialize our policies with the the supervised baseline and our value functions with the reward models. We set $\gamma = 1$ and $\lambda = 0.95$ for the advantage
estimation \cite{https://doi.org/10.48550/arxiv.1506.02438}, do 4 epochs of optimization with a batch size of 8 and run for 5,000 episodes. We set $w_l = 1$, $w_g = 1.5$ and $\beta = 0.05$ based on grid search among $\{0.05, 0.5, 1, 1.5, 2, 2.5\}$ for the full reward $R$. All experiments were performed on 8 NVIDIA V100 (32GB memory).

\subsection{Automatic Evaluation}
We first evaluated all the models with the widely used automatic metric ROUGE\cite{lin2004automatic} and reported ROUGE-1, ROUGE-2, and ROUGE-L in Table~\ref{Tab: rouge_results}. We found that
the performances were not better for synthesis data when there were less training data. When there was plenty of synthesis feedback, (\emph{HITL-synthesis with Full data}) can help improve over the supervised baseline, where the local reward was more important compared to the global reward. After incorporating ground-truth human feedback, our \emph{HITL-($r_g+r_l$)} model with both global and local rewards achieved the best performances even with less training data compared to synthesis baselines. The local rewards consistently brought in more performance boost because the conversation structural information in local rewards can help the systems more directly capture the important factors in the conversation. This indicates the effectiveness of our HITL framework for conversation summarization, as the human judgements were directly guiding the learning process. 

\subsection{Human Evaluation} \label{Sec:human}
Following \citet{https://doi.org/10.48550/arxiv.1909.01214} and \citet{https://doi.org/10.48550/arxiv.2009.01325}, we randomly sampled 200 conversations from the  DialogSum test set and asked annotators from Amazon Mechanical Turk to select the preferred summary from pairs of summaries generated by \emph{BART-large} and \emph{HITL-($r_g+r_l$)}. Turkers were asked to judge coherence, accuracy, coverage, and conciseness \footnote{The guidelines are the same with the guidelines for global feedback annotations described in the Appendix.}. To increase the annotation quality, we required Turkers to have a 98\%
approval rate and at least 10,000 approved tasks for their previous work. Each conversation was rated by three workers, and we used majority voting to decide the preferred summaries. The pay rate was 0.5\$ per hit. We measured the agreement by computing the Intra-Class Correlation (ICC) was 0.693, showing moderate agreement \cite{koo2016guideline}. 

\paragraph{Main Results} From Table~\ref{Tab: human_results}, we observed that summaries from our introduced (\emph{HITL-($r_g+r_l$)}) are much more preferred (favored in \textbf{82\%} cases) by human compared to supervised baseline (\emph{BART-large}). This significant improvements came from comparably \textit{small amount of annotations} (1000 dialogues). These indicated that the systems (\emph{HITL-($r_g+r_l$)}) that directly learn from small amount of global and local human feedback could generate higher-quality summaries with better human preferences compared to supervised baselines.

\paragraph{Evaluating the Global Reward Models} Based on the human preferences, we further examined the global reward model and compared it with its sub-dimensions as well as the ROUGE metric. Basically, we assume that the reward model agrees with human preferences when the model is assigning higher scores to these human-preferred summaries. As shown in Table~\ref{Tab:reward}, reward models learned from human  generally agree  well with human, where our global reward $r_g$ receives the highest agreement rate. This showed the high quality and effectiveness of our global feedback collection as well as the global reward models. As a result, our \emph{HITL-($r_g+r_l$)} model achieves better performances compared to baselines. 

\subsection{Generalization}
We then evaluated the generalization abilities of our \emph{HITL} summarization system and our learned global reward model $r_g$. We transferred the knowledge learned on DialogSum to another corpus, SAMSum \cite{gliwa-etal-2019-samsum} which summarize messenger-like conversations about daily topics, such as arranging meetings and discussing events. 

\paragraph{Generalization of HITL models} We first directly applied the whole \emph{HITL-($r_g+r_l$)} models trained on DialogSum to the SAMSum corpus. The results were visualized in Table~\ref{Tab:zero_shot}. The zero-shot evaluations on SAMSum got lower ROUGE scores compared to the models trained on SAMSum data, while our best model, \emph{HITL-($r_g+r_l$)}, achieved better performances compared to supervised baseline model (\emph{BART-large}). This showed that our policy empowered with human feedback can better generalize to other domains compared to supervised learning models, because our policy was learned from rewards that explicit indicated human preferrences. Such rewards are more general to different domains compared to supervised learning objectives which are specific to one dataset.


\paragraph{Generalization of the Global Reward Model} We then re-trained the \emph{HITL-($r_g$)} policy on SAMSum corpus while we directly utilized the global reward model $r_g(C,s)$ learned from human feedback on DialogSum data as the reward functions. We reported the results in Table~\ref{Tab:zero_shot} and observed that the \emph{HITL-($r_g$)} outperformed the supervised BART-large model on SAMSum in terms of ROUGE scores. This showed that our global reward models $r_g$ can be directly applied to other conversation summarization datasets to provide reinforcement learning rewards and boost performance because the global rewards learned from human feedback are implying the qualities of summaries in general rather than being limited to one specific domain.


\subsection{Ablation Study}
Here we performed two ablation studies to further study the impact of the quality and the quantity of human feedback in our HITL pipeline.

\paragraph{Perturbing the Qualities of Annotations} We compared \emph{HITL-($r_g+r_l$)} policy trained with annotations on the same 400 dialogues of three levels of qualities: (1) \textit{Synthesis Annotations} as described in Section~\ref{Sec:baseline}, (2) \textit{Noisy} which was the annotation from annotators without extensive training sessions, (3)  \textit{Clean} which was the annotation after the training sessions. We visualized the comparisons in Table~\ref{Tab:nosiy}, and found that the performances were better with higher quality annotations. This suggests that the quality of human feedback matters.

\paragraph{Increasing the Annotations} We then varied the number of annotations from 400 to 1000 in our \emph{HITL-($r_g+r_l$)} model in Table~\ref{Fig:nnumber}. The ROUGE scores were higher with more human annotations because of better reward learning and policy learning with more training data. This implies the importance of enough human feedback to learn and design better rewards.



\section{Conclusion}
In this work, we introduced two levels of conversation human feedback into the abstractive conversation summarization to generate human-preferred summaries. Specifically, we first collected the local and global human feedback to design the corresponding reward functions. We then learned the summarization policies via reinforcement learning to optimize the designed rewards. Extensive experiments in different settings and ablation studies on DialogSum and SAMSum corpus via both automatic and human evaluations demonstrated the effectiveness and generalization of our introduced HITL pipeline. For future work, we would like to explore incorporating human feedback in natural languages which are more general and explicit to indicate how to summarize conversations to improve the abstractive conversation summarization.


\bibliographystyle{acl_natbib}
\bibliography{anthology,acl2021}

\appendix

\section{Data Statistics for DialogueSUM and SAMSum}

\begin{table*}[t]
\center
\begin{tabular}{c|c|c|c|c} \toprule
\textbf{Dataset} & \textbf{\# of Dialogues}& \textbf{Avg \# of turns} & \textbf{Avg \# of Words} & \textbf{Avg Compression Rate} \\ \midrule \midrule

DialogSUM &13,406 &9.5 &131.0 &0.18 \\ \midrule
SAMSum &16,369 &11.1 &94.3 &0.30 \\

\bottomrule      
\end{tabular} \caption{Data Statistics of DialogSUM and SAMSum } \label{Tab:data_stats}
\end{table*}

\begin{figure*}[t!]
    \centering
    \begin{subfigure}[t]{1.0\textwidth}
           \centering
           \includegraphics[width=\textwidth]{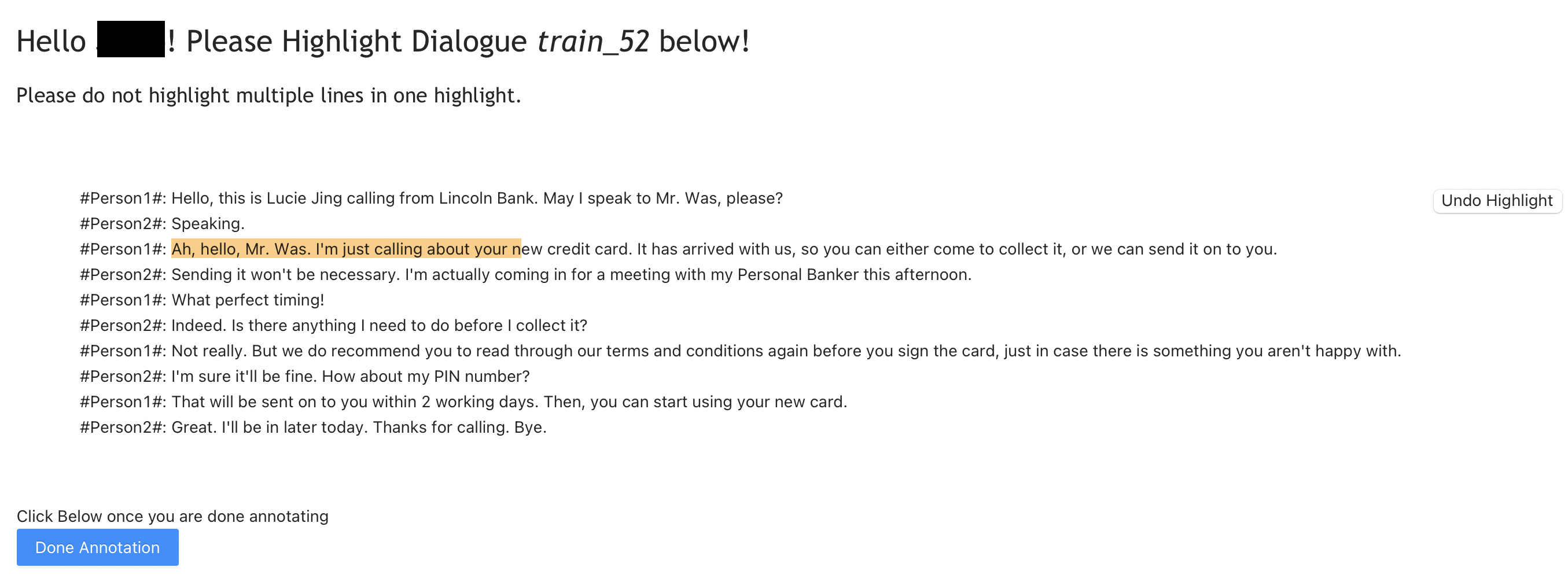}
            \caption{}
            \label{fig:a}
    \end{subfigure}
    \begin{subfigure}[t]{1.0\textwidth}
            \centering
            \includegraphics[width=\textwidth]{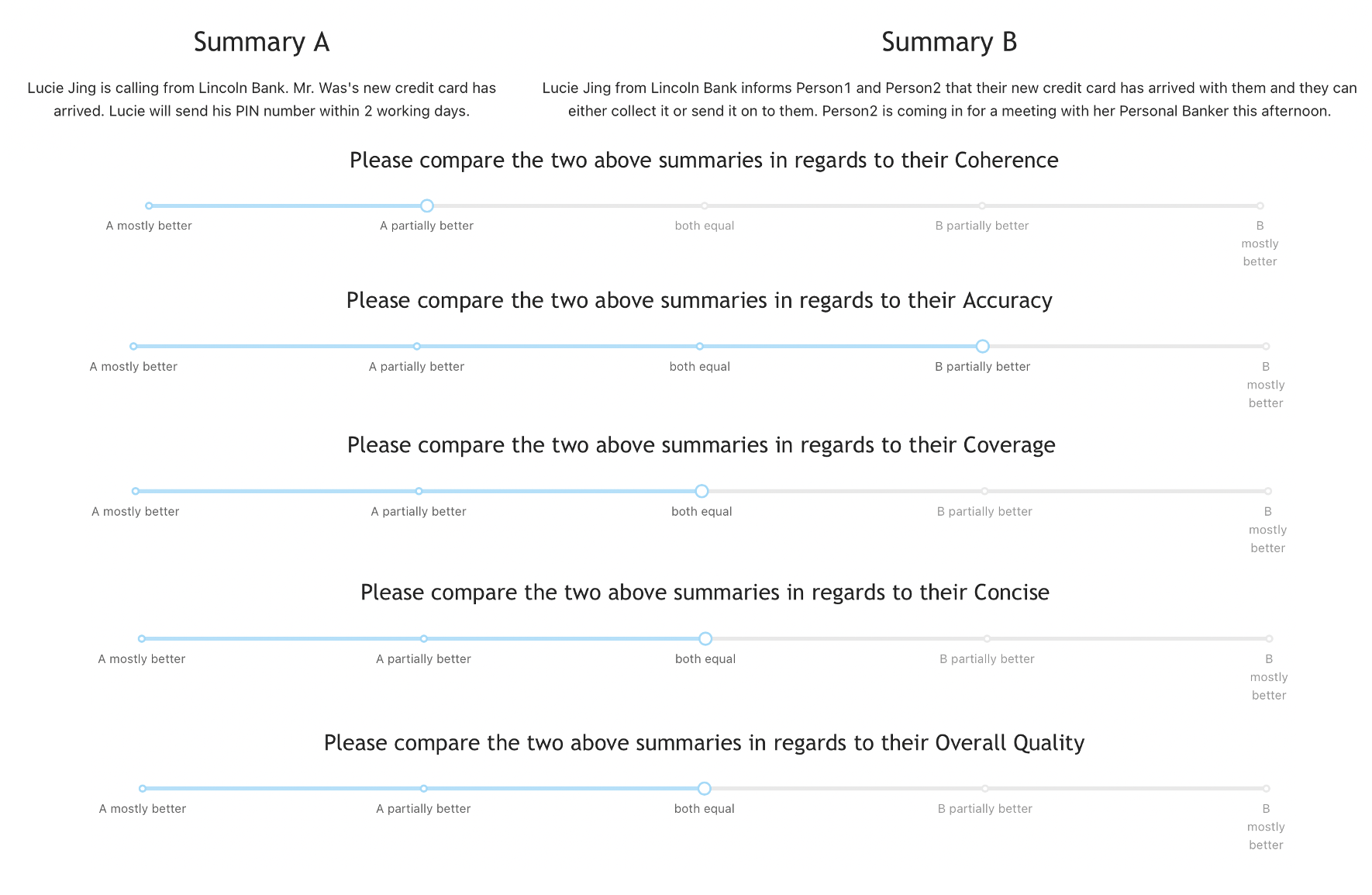}
            \caption{}
            \label{fig:b}
    \end{subfigure}
    \caption{The designed websits to collect data: (a) Highlighting key information in a given conversation. (b)  Comparing two given summaries in terms of given aspects.} \label{Fig:Interface}
\end{figure*}
\section{The Annotation Interface}
Since we hired and trained our own set of annotators, rather than using a crowd sourcing website such as Amazon Mechanical Turk, we built our own website to allow for a standardized, customized user interface for all annotators. The website contains the information for highlighting, summary comparisons as well as detailed instructions. From here we collect local and global guidance. For local guidance, we display one of the dialogues on the website. We ask the user to highlight salient information and then press next. Afterward, we display 3 pairs of summaries and ask the user to compare the pairs of summaries in 5 different dimensions. Screenshots from the website are shown in Figure~\ref{Fig:Interface}. Data collected from the website can be easily ported into a central database containing all of our human data.

\section{Global Feedback Guidelines}
We provide the annotators with 3 pairs of summaries sampled from the set of baseline summaries, and ask them to make comparisons in terms of \textit{Coherence, Accuracy, Coverage, Conciseness}, and \textit{Overall Quality}. For every comparison between summary A and summary B, the annotators need to grade upon a scale of 5 points: summary A mostly better, summary A partially better, equal, summary B partially better, summary B mostly better. We provide detailed guidelines to the annotators about those different dimensions:
\begin{itemize}
  \item \textbf{Coherence}: Summary is easy to understand and free of English errors. For comparing summaries against each other in Coherence, we ask the annotators to compare the number and severity of grammatical, syntax, and spelling errors of each summary against each other.

  \item \textbf{Accuracy}: Information that stated in the summary is accurate and does not incorrect information. Summary is not misleading and has too much errors. For comparing summaries against each other in Accuracy, we ask the annotators discover the amount and severity of inaccurate statements that occur in the summaries against each other.
    
  \item \textbf{Coverage}: Mentions main information of the conversations. It conveys the most salient information from the dialogue. For comparing summaries against each other in Coverage, we ask the annotators to look at the number of events in each summary. Also taking into the factor of importance of events, we ask the annotator to compare the number of events against the pair of summaries.
  
  \item \textbf{Conciseness}: Summary is short and to the point. It does not have too much unimportant information that is not included in the salient information. For comparing summaries against each other in Conciseness, we ask the annotators to mainly look at the length of the summaries. Then we check if any information doesn't fit, and penalize as such.

 \item \textbf{Overall Quality}: We ask the annotator to use all of the above information and other related context to give an overall rating. Even though we asked the annotator to consider all the information, we asked the annotator to factor coverage and accuracy more into their decision for Overall Quality. This is because it is of at most importance for a dialogue summary to accurately summarize the salient information of the dialogue.

\end{itemize}

\end{document}